\documentclass[conference,english,]{IEEEtran}
\usepackage{lmodern}
\usepackage{amssymb,amsmath}
\usepackage{ifxetex,ifluatex}
\ifnum 0\ifxetex 1\fi\ifluatex 1\fi=0 
  \usepackage[T1]{fontenc}
  \usepackage[utf8]{inputenc}
\else 
  \ifxetex
    \usepackage{mathspec}
  \else
    \usepackage{fontspec}
  \fi
  \defaultfontfeatures{Ligatures=TeX,Scale=MatchLowercase}
\fi
\IfFileExists{upquote.sty}{\usepackage{upquote}}{}
\IfFileExists{microtype.sty}{%
\usepackage{microtype}
\UseMicrotypeSet[protrusion]{basicmath} 
}{}
\usepackage[unicode=true]{hyperref}
\hypersetup{
            pdftitle={App for Resume-Based Job Matching with Speech Interviews and Grammar Analysis: A Review},
            pdfkeywords={machine learning, deep learning, natural
language processing, speech recognition},
            pdfborder={0 0 0},
            breaklinks=true}
\ifnum 0\ifxetex 1\fi\ifluatex 1\fi=0 
  \usepackage[shorthands=off,main=english]{babel}
\else
  \usepackage{polyglossia}
  \setmainlanguage[]{}
\fi
\usepackage[square,sort,comma,numbers]{natbib}
\usepackage{longtable,booktabs}
\usepackage{supertabular}

\IfFileExists{footnote.sty}{\usepackage{footnote}\makesavenoteenv{long table}}{}
\usepackage{graphicx,grffile}
\makeatletter
\def\maxwidth{\ifdim\Gin@nat@width>\linewidth\linewidth\else\Gin@nat@width\fi}
\def\maxheight{\ifdim\Gin@nat@height>\textheight\textheight\else\Gin@nat@height\fi}
\makeatother
\setkeys{Gin}{width=\maxwidth,height=\maxheight,keepaspectratio}
\IfFileExists{parskip.sty}{%
\usepackage{parskip}
}{
\setlength{\parindent}{0pt}
\setlength{\parskip}{6pt plus 2pt minus 1pt}
}
\ifx\paragraph\undefined\else
\let\oldparagraph\paragraph
\renewcommand{\paragraph}[1]{\oldparagraph{#1}\mbox{}}
\fi
\ifx\subparagraph\undefined\else
\let\oldsubparagraph\subparagraph
\renewcommand{\subparagraph}[1]{\oldsubparagraph{#1}\mbox{}}
\fi

\makeatletter
\def\fps@figure{htbp}
\makeatother

\makeatletter
\@ifpackageloaded{subfig}{}{\usepackage{subfig}}
\@ifpackageloaded{caption}{}{\usepackage{caption}}
\AtBeginDocument{%

}
\AtBeginDocument{%

}
\newcounter{pandoccrossref@subfigures@footnote@counter}
{\end{figure}%
\addtocounter{footnote}{-\value{pandoccrossref@subfigures@footnote@counter}}
\@for\f:=\global@pandoccrossref@subfigures@footnotes\do{\stepcounter{footnote}\footnotetext{\f}}%
\gdef\global@pandoccrossref@subfigures@footnotes{}}
\@ifpackageloaded{float}{}{\usepackage{float}}
\floatstyle{ruled}
\@ifundefined{c@chapter}{\newfloat{codelisting}{h}{lop}}{\newfloat{codelisting}{h}{lop}[chapter]}
\floatname{codelisting}{Listing}

\makeatother

\title{App for Resume-Based Job Matching with Speech Interviews and
Grammar Analysis: A Review}

\author{
\IEEEauthorblockN{Tanmay Kulkarni\IEEEauthorrefmark{1}, Yuvraj
Pardeshi\IEEEauthorrefmark{1}, Yash Shah\IEEEauthorrefmark{1}, Vaishnvi
Sakat\IEEEauthorrefmark{1} and}
\IEEEauthorblockN{Sapana Bhirud\IEEEauthorrefmark{2}}
\IEEEauthorblockA{%
\\
\IEEEauthorrefmark{1}Department of Artificial Intelligence and Machine
Learning, P.E.S. Modern College of Engineering, India\\
\IEEEauthorrefmark{2}Professor, Department of Artificial Intelligence
and Machine Learning, P.E.S. Modern College of Engineering, India\\
\{
\href{mailto:tanmay\_kulkarni@moderncoe.edu.in}{tanmay\_kulkarni}
,
\href{mailto:yuvraj\_pardeshi@moderncoe.edu.in}{yuvraj\_pardeshi}
,
\href{mailto:yash\_shah@moderncoe.edu.in}{yash\_shah}
,
\href{mailto:vaishnavi\_sakat@moderncoe.edu.in}{vaishnavi\_sakat}
,
\href{mailto:sapana.bhirud@moderncoe.edu.in}{sapana.bhirud}
\} @moderncoe.edu.in}
}

\date{\today}

\begin{document}
\maketitle
\begin{abstract}
Through the advancement in natural language processing (NLP),
specifically in speech recognition, fully automated complex systems
functioning on voice input have started proliferating in areas such as
home automation. These systems have been termed Automatic Speech
Recognition Systems (ASR). In this review paper, we explore the
feasibility of an end-to-end system providing speech and text based
natural language processing for job interview preparation as well as
recommendation of relevant job postings. We also explore existing
recommender-based systems and note their limitations. This literature
review would help us identify the approaches and limitations of the
various similar use-cases of NLP technology for our upcoming project.
\end{abstract}


\section{Introduction}\label{introduction}

~~~~The process of finding a job has undergone significant change due to
the quick improvements in technology and changing industry demands.
Recognizing the changing environment, our project aims to close the gap
between educational institutions and business needs. We think that
preparing for placements entails more than just knowing how to respond
to interview questions; it also entails being aware of the industry, the
labor market, and the abilities that make a person a priceless asset to
any firm.

~~~~Navigating the challenging environment of job preparation and
placement can be challenging in an era of ferocious competition. The
modern job market takes more than simply a degree; it also demands a
calculated strategy, flawless preparation, and a thorough knowledge of
industry dynamics. Our project called `CareerSpeak: The Placement
Preparation App' aims to arm ambitious professionals with the
information, abilities, and self-assurance they need to succeed in their
professional endeavors.

\section{Literature Survey}\label{literature-survey}

\subsection{\texorpdfstring{Resume Parser Using Natural Language
Processing Techniques
\citep{resume_parser_nlp}}{Resume Parser Using Natural Language Processing Techniques {[}@resume\_parser\_nlp{]}}}\label{resume-parser-using-natural-language-processing-techniques-resume_parser_nlp}

~~~~The paper proposes a model that uses Natural Language Processing
(NLP) techniques to extract details and statistics from resumes and rank
them based on company preferences and requirements. The model aims to
build a job portal where employees and applicants can upload their
resumes for specific jobs. The NLP technique is used to parse the
necessary information and generate structured resumes. Resumes are also
ranked based on the company's skill requirements and the skills
mentioned by the applicants. Techniques such as neural networks, CRF,
CNN, and segmentation models are used for information extraction from
resumes.

~~~~The results of the system involve parsing resumes into plain
documents, extracting entities, and comparing them with required
keywords. The results are presented in the form of pie charts and bar
graphs.

\subsection{\texorpdfstring{A Keyword Extraction Method Based on
learning to Rank
\citep{keyword_extraction_rank}}{A Keyword Extraction Method Based on learning to Rank {[}@keyword\_extraction\_rank{]}}}\label{a-keyword-extraction-method-based-on-learning-to-rank-keyword_extraction_rank}

~~~~This paper speaks about TransR method for knowledge graph
completion. TransR is an approach that combines graph embedding and rule
mining techniques to improve the accuracy of knowledge graph completion.
It incorporates both entity and relation embeddings to enhance the
performance of link prediction and triple classification tasks.
Experimental results on benchmark datasets demonstrate the effectiveness
of the TransR approach, outperforming existing methods in terms of
evaluation metrics such as mean reciprocal rank and precision at
different ranks.

~~~~The paper also discusses the limitations of the proposed method and
suggests future research directions in the field of knowledge graph
completion.

\subsection{\texorpdfstring{Automatic Extraction of Usable Information
from Unstructured Resumes to Aid Search
\citep{automatic_extraction}}{Automatic Extraction of Usable Information from Unstructured Resumes to Aid Search {[}@automatic\_extraction{]}}}\label{automatic-extraction-of-usable-information-from-unstructured-resumes-to-aid-search-automatic_extraction}

~~~~The paper proposes a system for automated resume information
extraction using natural language processing (NLP) techniques to support
rapid resume search and management. The system is capable of extracting
several important fields from free format resumes, including personal
information, education, contact telephone number, postal address,
languages known, present company, and designation. The proposed system
can handle a large variety of resumes in different document formats with
a precision of 91\% and a recall of 88\%.

~~~~The system aims to eliminate the need for job seekers to fill in
predefined templates and allows enterprises to extract the required
information from any format of resume automatically. The paper
highlights the challenges of extracting information from
non-standardized resume structures and emphasizes the benefits of an
automated system for resume management, including the construction of an
electronic resume database and quick processing of resumes. The
performance of the system is evaluated using precision and recall
metrics on a set of resumes that were not used as reference resumes to
build the knowledge base.

\subsection{\texorpdfstring{Overview of the Speech Recognition
Technology
\citep{speech_recognition_overview}}{Overview of the Speech Recognition Technology {[}@speech\_recognition\_overview{]}}}\label{overview-of-the-speech-recognition-technology-speech_recognition_overview}

~~~~The paper highlights two key approaches in speech recognition:
Hidden Markov Model (HMM) and Artificial Neural Network (ANN). HMM is a
statistical model used for fast and accurate speech recognition, while
ANN mimics biological nervous systems and offers features like training,
parallel processing, rapid judgment, and fault tolerance. Artificial
neural networks (ANN) are employed to improve the adaptability and
response of speech recognition systems to error inputs. Hidden Markov
Models (HMM) are utilized as a statistical model to train the acoustic
and voice models in speech recognition systems, leading to accurate and
fast recognition results.

~~~~The paper addresses challenges in noisy environments, such as
variations in pronunciation, speech rate, pitch, and formant changes and
suggests the use of new signal analysis and processing approaches.
Additionally Representative speech recognition methods, including
dynamic time warping (DTW), vector quantization (VQ), and support vector
machine (SVM), are also mentioned in the paper, but the focus is on HMM
and ANN methods.

\section{Methodology}\label{methodology}

~~~~The system consists of 3 parts- frontend, middleware and backend.
The user interacts with the frontend, which in turn interacts with the
middleware to process data and transfer information between the user in
the frontend and the machine learning models in the backend.

\begin{figure}
\centering
\includegraphics{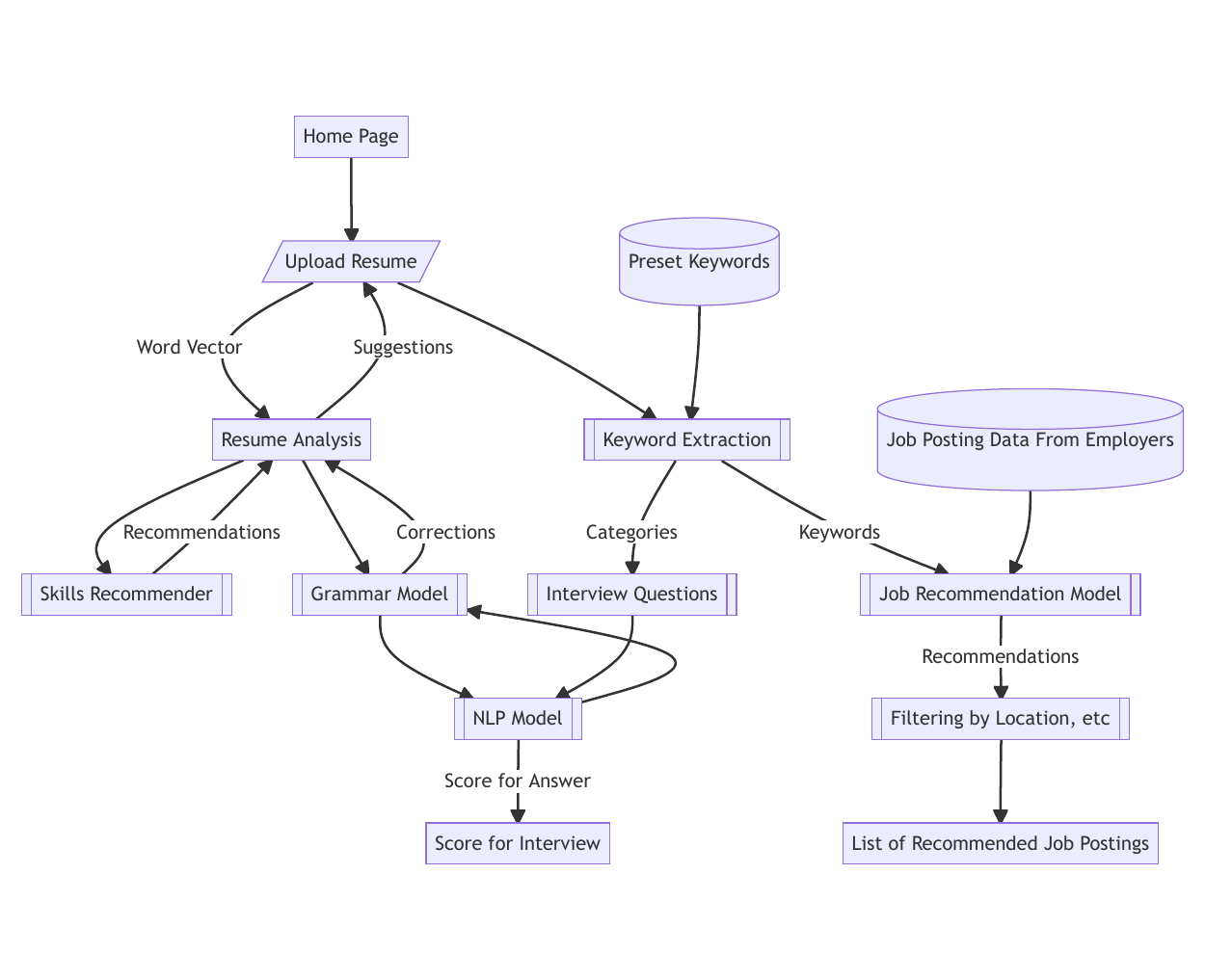}
\caption{Architecture of Proposed System}\label{fig:architecture}
\end{figure}

~~~~The system is divided into 3 modules- Resume Parser module, Mock
Interviewer module and Job Recommender module. These modules interact to
parse the resume uploaded by the user, extracting the important keywords
from the resume, and generating pertinent questions based on the
extracted keywords. This system architecture is visualized in
fig.~\ref{fig:architecture}

\begin{enumerate}
\def\labelenumi{\arabic{enumi}.}
\item
  Resume Parser:

  ~~~~When the user uploads a resume, the pdf document is parsed into
  text. This parsed data is stored as a string variable in memory
  \citep{resume_parser_nlp} \citep{keyword_extraction_rank}, persisting
  until the user exits the app. The user may choose to have the resume
  persist as a pdf document, which would be parsed again as required.
  The whole parsed resume data is given to LanguageTool for grammatical
  processing. LanguageTool is a Java-based grammar checker that filters
  the given input text based on predefined rules. These rules can be
  extended or selectively removed for specific applications.

  ~~~~LanguageTool provides the matching rule and a suggestion for the
  rule along with the line and column where the rule match occurs. This
  output is extracted into the middleware and then visually shown to the
  user from the frontend with a line underlining the characters at the
  line and column, a textbox showing the rule matched (or a simplified
  message) and the suggestion.

  \begin{figure}
  \centering
  \includegraphics[width=\textwidth,height=0.25\textheight]{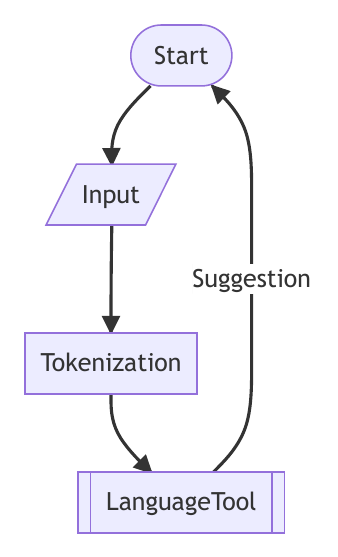}
  \caption{Flowchart for Grammar Checking
  Module}\label{fig:languagetool}
  \end{figure}

  ~~~~As shown fig.~\ref{fig:languagetool}, this process loops at each
  update the user makes to the text to provide real-time
  grammar-checking.\\
\item
  Mock Interviewer:

  ~~~~This module uses Automatic Speech Recognition (ASR).
  \citep{speech_recognition_overview} \citep{speech_to_text}
  \citep{automatic_extraction}

  ~~~~It uses the WhisperAI Speech to Text pretrained Deep Learning
  model \citep{automatic_extraction} from OpenAI to perform analysis on
  the sentences enunciated by the user to transcribe it into text.

  ~~~~In the interviewer module, a NoSQL database containing predefined
  general questions. These would be such that they would apply to all
  candidates, irrespective of their technical background. Some query
  processing may also be done to filter the questions based on the
  user's employment history.

  ~~~~The keywords extracted \citep{resume_parser_nlp}
  \citep{keyword_extraction_rank} from the resume play a vital role in
  fetching questions relating to technologies relevant to the
  educational and employment history of the user. This will test the
  technical knowledge of the user. The questions for each keyword or
  each set of keywords need to be designed manually, so we are limiting
  them to the fields of Web Development, Machine Learning and Data
  Science. More categories and questions can be easily added as needed
  due to the modular nature of the NoSQL database and the keyword
  extractor.

  ~~~~The questions also have ideal answers linked to them so that the
  user can get instant feedback of where the interviewing module found
  their answer to be lacking. This will also make the module more
  transparent in its scoring of interviews. Interviews will get a score
  indicating how close to ideal their answers were. More conditions such
  as avoiding unclear enunciations\citep{speech_to_text}, etc. may be
  added as the module is developed.\\
\item
  Job Recommender:

  ~~~~This module is also refered to as Job Recommender System (JRS).
  \citep{review_job_recommender}

  ~~~~The keywords extracted from the resume are used along with the
  professional and educational qualification fields by the recommender
  model to provide a list of potentially relevant job opportunities.

  ~~~~The job listings may be added by the employers individually or a
  web scraping tool \citep{job_recommender_skills} may be used for
  finding the listings. These listings will be stored in a RDBMS so that
  they can be retrieved by simple SQL queries according to the
  categories \citep{resume_parser_nlp} found from the keywords.

  ~~~~The final list of postings will be forwarded to the frontend,
  where the user may filter them based on location, salary, and other
  temporal metrics.
\end{enumerate}

\section{Limitations}\label{limitations}

\subsection{\texorpdfstring{Job Recommender Systems: A Review
\citep{review_job_recommender}}{Job Recommender Systems: A Review {[}@review\_job\_recommender{]}}}\label{job-recommender-systems-a-review-review_job_recommender}

~~~~The classification of hybrid methods in the job recommender system
(JRS) literature may still have some overlap and similarity between
different classes, which could lead to confusion in understanding the
methods used.

\subsection{\texorpdfstring{Resume Parser Using Natural Language
Processing Techniques
\citep{resume_parser_nlp}}{Resume Parser Using Natural Language Processing Techniques {[}@resume\_parser\_nlp{]}}}\label{resume-parser-using-natural-language-processing-techniques-resume_parser_nlp-1}

~~~~The paper does not address potential biases or limitations in the
ranking algorithm used to prioritize resumes based on company
preferences and requirements. Ensuring fairness and avoiding bias in the
ranking process is crucial.

\subsection{\texorpdfstring{Job Recommendation based on Job Seeker
Skills: An Empirical Study
\citep{job_recommender_skills}}{Job Recommendation based on Job Seeker Skills: An Empirical Study {[}@job\_recommender\_skills{]}}}\label{job-recommendation-based-on-job-seeker-skills-an-empirical-study-job_recommender_skills}

~~~~The paper does not explore the scalability or efficiency of the
proposed framework, which could be important considerations for
real-world application design.

\subsection{\texorpdfstring{VOICE RECOGNITION SYSTEM: SPEECH-TO-TEXT
\citep{speech_to_text}}{VOICE RECOGNITION SYSTEM: SPEECH-TO-TEXT {[}@speech\_to\_text{]}}}\label{voice-recognition-system-speech-to-text-speech_to_text}

~~~~The paper does not provide any information about the limitations of
the adapted feature extraction technique or the speech recognition
approach used in the system. The paper does not discuss any limitations
of the low pass filter with finite impulse response or the performance
evaluation at signal to noise ratio level.

\subsection{\texorpdfstring{Overview of the Speech Recognition
Technology
\citep{speech_recognition_overview}}{Overview of the Speech Recognition Technology {[}@speech\_recognition\_overview{]}}}\label{overview-of-the-speech-recognition-technology-speech_recognition_overview-1}

~~~~The paper highlights the poor adaptability of speech recognition
systems, as they are highly dependent on the environment in which the
speech training data is collected. This limits their performance in
different environments.

\section{Conclusion}\label{conclusion}

~~~~We have reviewed various use-cases and implementations of natural
language processing and keyword extraction in resumes. This review has
shown that a system for job recommendation based on keyword detection
and mock interview based on automatic speech recognition is feasible. We
have also determined the scope for our proposed system, limiting to
three job postings and only one language. The approaches to keyword
extraction and automatic speech recognition explored in this review show
the modularity and flexibily of this technology to adapt it to other
domains and use-cases.

~~~~Our upcoming research paper based on our implementation of the
proposed end-to-end job recommender, resume grammar-checker and mock
interviewer app will be utilizing these technologies for easing the
placement process.

\renewcommand\refname{References}
\bibliography{content/bibliography.bib}

\begin{thebibliography}{1}
\providecommand{\url}[1]{#1}
\csname url@samestyle\endcsname
\providecommand{\newblock}{\relax}
\providecommand{\bibinfo}[2]{#2}
\providecommand{\BIBentrySTDinterwordspacing}{\spaceskip=0pt\relax}
\providecommand{\BIBentryALTinterwordstretchfactor}{4}
\providecommand{\BIBentryALTinterwordspacing}{\spaceskip=\fontdimen2\font plus
\BIBentryALTinterwordstretchfactor\fontdimen3\font minus
  \fontdimen4\font\relax}
\providecommand{\BIBforeignlanguage}[2]{{%
\expandafter\ifx\csname l@#1\endcsname\relax
\typeout{** WARNING: IEEEtran.bst: No hyphenation pattern has been}%
\typeout{** loaded for the language `#1'. Using the pattern for}%
\typeout{** the default language instead.}%
\else
\language=\csname l@#1\endcsname
\fi
#2}}
\providecommand{\BIBdecl}{\relax}
\BIBdecl

\bibitem{resume_parser_nlp}
S.~Bhor, V.~Gupta, V.~Nair, H.~Shinde, and M.~S. Kulkarni, ``Resume parser
  using natural language processing techniques,'' \emph{Int J Res Eng Sci
  (IJRES)}, vol.~9, no.~6, pp. 01--06, 2021.

\bibitem{keyword_extraction_rank}
X.~Cai and S.~Cao, ``A keyword extraction method based on learning to rank,''
  in \emph{2017 13th International Conference on Semantics, Knowledge and Grids
  (SKG)}, 2017, pp. 194--197.

\bibitem{automatic_extraction}
R.~Singh, H.~Yadav, M.~Sharma, S.~Gosain, and R.~R. Shah, ``Automatic speech
  recognition for real time systems,'' in \emph{2019 IEEE Fifth International
  Conference on Multimedia Big Data (BigMM)}, 2019, pp. 189--198.

\bibitem{speech_recognition_overview}
J.~Meng, J.~Zhang, and H.~Zhao, ``Overview of the speech recognition
  technology,'' in \emph{2012 Fourth International Conference on Computational
  and Information Sciences}, 2012, pp. 199--202.

\bibitem{speech_to_text}
P.~Das, K.~Acharjee, P.~Das, and V.~Prasad, ``Voice recognition system:
  Speech-to-text,'' \emph{Journal of Applied and Fundamental Sciences}, vol.~1,
  pp. 2395--5562, 11 2015.

\bibitem{review_job_recommender}
C.~de~Ruijt and S.~Bhulai, ``Job recommender systems: A review,'' 2021.

\bibitem{job_recommender_skills}
J.~C. Valverde-Rebaza, R.~Puma, P.~Bustios, and N.~C. Silva, ``Job
  recommendation based on job seeker skills: An empirical study.'' in
  \emph{Text2Story@ ECIR}, 2018, pp. 47--51.

\end{thebibliography}

\end{document}